\documentclass[10pt]{article}

\usepackage[margin=1in]{geometry}
\usepackage{amsmath,amssymb,amsthm}
\usepackage{graphicx}
\usepackage{booktabs}
\usepackage{multirow}
\usepackage{microtype}
\usepackage[numbers,sort&compress]{natbib}
\usepackage{xcolor}
\usepackage[colorlinks=true,linkcolor=blue!60!black,citecolor=blue!60!black,urlcolor=blue!60!black]{hyperref}
\usepackage{caption}
\newcommand{\loss}{\mathcal{L}}
\newcommand{\R}{\mathbb{R}}
\newcommand{\E}{\mathbb{E}}

\title{Level-Crossing Density as a Mesh-Free High-Frequency\\
Auxiliary Loss for Implicit Neural Representations}

\author{%
Gunner Levi Howe\\
Independent Researcher\\
\texttt{gunnerlevihowe@gmail.com}%
}

\date{July 2026}

\begin{document}
\maketitle

\begin{abstract}
Coordinate-MLP implicit neural representations (INRs) exhibit \emph{spectral bias}: they
fit low-frequency structure quickly and high-frequency detail slowly or not at all.
Existing remedies act in the frequency domain (e.g., the Focal Frequency Loss), on
activations and encodings (SIREN, Fourier features, FINER), or through coarse-to-fine
curricula---and the loss-based remedies presuppose a regular sampling grid on which a
discrete Fourier transform is available. We revisit a classical object from random-field
theory, the \emph{Rice level-crossing density}, and turn it into a differentiable
training objective. By the co-area formula, the density of level crossings of a field is
a direct functional of its gradient magnitude concentrated on level sets, and for
stationary processes it is monotone in the ratio of spectral moments
$\sqrt{\lambda_2/\lambda_0}$---i.e., it is a spatial-domain, closed-form proxy for
high-frequency content. We construct a smoothed Monte-Carlo estimator of the crossing
density that is evaluated pointwise at arbitrary sample locations, requires no grid, no
FFT, and no mesh, and match it to the empirical crossing density of the supervision
signal across a set of levels. We validate the estimator against exact crossing counts
and the Rice formula, then evaluate the loss under strict information parity against
the Focal Frequency Loss, normalized Sobolev supervision, and MSE-only training, on
regular grids and on scattered non-uniform samples, and we report the outcome without
embellishment. Where supervision is scarce and irregular, \emph{every} auxiliary
spectral loss is transformative ($+2.3$--$3.0$\,dB over MSE-only on PE-MLP,
$+1.4$--$1.8$\,dB on SIREN); on an edge-dominated natural image the crossing-density
loss matches the alternatives---within $0.2$--$0.5$\,dB of FFL on PSNR, matching or
exceeding it on SSIM---without beating them, while on a statistically homogeneous
texture, the regime closest to the stationary fields of the Rice theory, it overtakes
FFL by $0.6$\,dB. Its further distinctions are structural:
it is the only loss requiring neither a sampling grid nor trustworthy pointwise
gradient targets, and an oracle-gradient diagnostic shows it is uniquely insensitive
to gradient-target quality. On dense regular grids, where auxiliary drives are
unnecessary, finite-difference gradient targets actively hurt---a caution that applies
equally to Sobolev training. We release code, raw results, and all experiment
scripts.
\end{abstract}

\section{Introduction}
\label{sec:intro}

Implicit neural representations (INRs) parameterize a signal as a neural network
$f_\theta:\R^d\to\R^m$ mapping coordinates to values, and have become a standard
representation for images, signed distance fields, and radiance fields
\citep{sitzmann2020siren,park2019deepsdf,mildenhall2020nerf,xie2022neuralfields}. Their
best-known failure mode is \emph{spectral bias}
\citep{rahaman2019spectral,basri2020frequency}: trained with a plain reconstruction
loss, coordinate MLPs learn low frequencies first and high frequencies slowly, blurring
fine detail. The community's responses fall into three families: (i)~\emph{architectural}
fixes---Fourier feature encodings \citep{tancik2020fourier}, periodic activations
\citep{sitzmann2020siren}, their variable-periodic \citep{liu2024finer}, wavelet
\citep{saragadam2023wire}, and Gaussian \citep{ramasinghe2022beyond} relatives,
multiplicative filter networks \citep{fathony2021mfn}, band-limited architectures
\citep{lindell2022bacon}, and hash grids \citep{muller2022instant};
(ii)~\emph{loss-based} fixes---frequency-domain objectives such as the Focal Frequency
Loss (FFL) \citep{jiang2021ffl} and gradient-domain (Sobolev) supervision
\citep{czarnecki2017sobolev,yuan2022sobolev}; and (iii)~\emph{curricula} that schedule
frequency content coarse-to-fine \citep{hertz2021sape,yang2023freenerf}.

The loss-based family has a structural blind spot. A discrete Fourier transform is
defined on a regular grid. When supervision arrives as scattered samples---point clouds
for neural SDFs, non-uniformly sampled sensor data, adaptively sampled renderings---a
frequency-domain loss must first \emph{resample} the scattered data onto a grid, and the
interpolation step smears exactly the high-frequency content the loss is meant to
supervise. Gradient-matching losses are pointwise and thus mesh-free, but they supervise
the gradient \emph{at each point}, so their targets inherit, pointwise and at full
strength, whatever noise the gradient-estimation step introduces.

This paper proposes a third kind of spatial-domain objective, borrowed from random-field
theory. The \emph{Rice formula} \citep{rice1944,kac1943} states that for a stationary
Gaussian process the expected number of crossings of level $u$ per unit length is
\begin{equation}
\mu(u) \;=\; \frac{1}{\pi}\sqrt{\frac{\lambda_2}{\lambda_0}}\;
e^{-u^2/(2\lambda_0)},
\qquad \lambda_k = \int \omega^k \, dS(\omega),
\label{eq:rice}
\end{equation}
where $S$ is the spectral measure. The crossing rate is monotone in
$\sqrt{\lambda_2/\lambda_0}$, the RMS angular frequency: \emph{how often a field crosses
a level is a direct, closed-form functional of how much high-frequency energy it
carries}. This link between zero-crossing rates and spectral content is classical in
signal processing \citep{kedem1986zerocrossings,longuethiggins1957}; our contribution is
not the link itself but its \emph{instantiation as a differentiable training objective
for INRs}, and an empirical study of where such an objective helps.

Concretely, we estimate the crossing density of the INR's output field with a smoothed
Monte-Carlo form of the Kac--Rice integrand,
\begin{equation}
\hat c_\varepsilon(u) \;=\; \frac{1}{N}\sum_{i=1}^N
\delta_\varepsilon\!\bigl(f_\theta(x_i)-u\bigr)\,
\bigl\lVert \nabla_x f_\theta(x_i)\bigr\rVert,
\label{eq:estimator}
\end{equation}
where the $x_i$ are the (arbitrary, possibly scattered) training coordinates,
$\delta_\varepsilon$ is a Gaussian bump, and $\nabla_x f_\theta$ is exact via automatic
differentiation. By the co-area formula, \eqref{eq:estimator} estimates crossings per
unit length in 1D, level-set length per unit area in 2D, and level-set area per unit
volume in 3D. The auxiliary loss matches $\hat c_\varepsilon(u)$ to the \emph{empirical}
crossing density of the ground truth at a small set of levels. Everything is pointwise:
no grid, no FFT, no mesh, no stationarity or Gaussianity assumption.

Because a crossing-density profile is a \emph{distributional} statistic---an average
over the whole batch---it discards phase information: it says how much level-set
geometry the field should have, not where. It is therefore an auxiliary drive alongside
a reconstruction loss (which anchors locations), not a stand-alone objective, and its
distributional character is exactly what we expect to make it robust on scattered
domains: pointwise errors in derived gradient targets average out in
\eqref{eq:estimator} instead of being enforced point by point.

\paragraph{Claims.} We make three claims, each tested in
Section~\ref{sec:experiments}. (1)~The estimator \eqref{eq:estimator} is an accurate,
differentiable crossing-density estimator (validated against exact counts and against
\eqref{eq:rice}). (2)~As an auxiliary loss it is a \emph{working} high-frequency
drive: wherever supervision is scarce it recovers most of the gap between MSE-only
training and the information ceiling set by the samples, on two different backbones.
(3)~Against the strongest loss-based alternatives under strict information parity, it
achieves PSNR parity on edge-dominated natural-image content (with SSIM matching or
exceeding FFL's and unique robustness to gradient-target quality) and overtakes FFL only on
statistically homogeneous texture---the regime the Rice theory actually describes. We
consider the disciplined measurement of (3), including its negative component, a
contribution: the hypothesis that crossing statistics would broadly \emph{beat}
resampled frequency losses off-grid was plausible, cheap to state, and is here tested
and bounded.

\paragraph{What we do not claim.} Coarse-to-fine frequency curricula exist
\citep{hertz2021sape,yang2023freenerf} and we do not claim one. The
crossing-rate--frequency link is textbook \citep{kedem1986zerocrossings}; we claim only
the differentiable-loss instantiation and its evaluation. On regular grids we expect,
and report, rough parity rather than dominance.

\section{Related work}
\label{sec:related}

\paragraph{Spectral bias and architectural remedies.}
Spectral bias was characterized by \citet{rahaman2019spectral}, with
\citet{basri2020frequency} extending the analysis to non-uniform input densities---%
directly relevant to our scattered-sampling setting, since their theory predicts that
convergence at a given frequency slows further in sparsely sampled regions. Fourier
feature encodings \citep{tancik2020fourier} and periodic activations
\citep{sitzmann2020siren} move the representable band up; FINER \citep{liu2024finer}
makes the supported band tunable via variable-periodic activations; WIRE
\citep{saragadam2023wire}, Gaussian activations \citep{ramasinghe2022beyond},
multiplicative filter networks \citep{fathony2021mfn}, BACON \citep{lindell2022bacon},
and hash encodings \citep{muller2022instant} are alternative routes. All of these change
the \emph{architecture}; our proposal is orthogonal and combines with them
(Section~\ref{sec:composability}).

\paragraph{Loss-based remedies.}
The Focal Frequency Loss \citep{jiang2021ffl} compares DFT spectra of prediction and
target with an adaptive per-frequency weight and is the standard frequency-domain
objective; it requires a regular grid. Sobolev training supervises derivatives
\citep{czarnecki2017sobolev}, applied to INRs with finite-difference image derivatives
by \citet{yuan2022sobolev}; it is mesh-free but pointwise. Eikonal-type losses in shape
representation \citep{gropp2020igr} likewise constrain $\lVert\nabla f\rVert$ pointwise
(to unit norm for SDFs); our loss instead constrains an integral functional of
$\lVert\nabla f\rVert$ restricted to level sets. Frequency curricula
\citep{hertz2021sape,yang2023freenerf} schedule the encoding rather than add an
objective.

\paragraph{Level crossings and the Kac--Rice formula.}
Crossing counts of random processes go back to \citet{kac1943} and \citet{rice1944};
modern treatments include \citet{adler2007random,azais2009level}, and
\citet{longuethiggins1957} developed the 2D (level-set length) theory for random
surfaces. Zero-crossing rates as frequency features are classical in signal processing
and time-series discrimination \citep{kedem1986zerocrossings}. To our knowledge---%
including searches over arXiv abstracts and the recent INR literature---level-crossing
densities have not previously been used as a training objective for neural fields; the
Kac--Rice machinery appears in machine learning mainly for counting critical points of
loss landscapes, a different use entirely.

\section{Method}
\label{sec:method}

\subsection{Background: crossings, spectra, and the co-area identity}
\label{sec:background}

\begin{figure}[t]
\centering
\includegraphics[width=\linewidth]{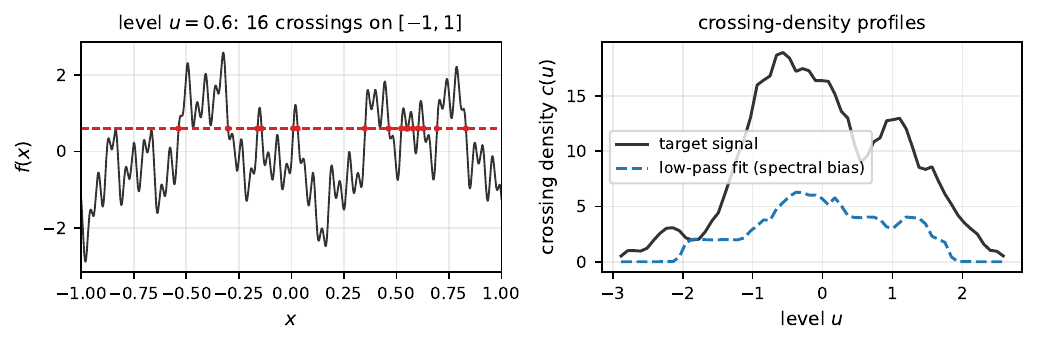}
\caption{\textbf{Left:} a multisine signal, one level $u$, and its crossings; the
crossing density at level $u$ is the number of such intersections per unit length.
\textbf{Right:} crossing-density profiles $c(u)$ of the full signal versus its low-pass
approximation---the profile collapses when high frequencies are missing, which is the
signal the loss exploits.}
\label{fig:method}
\end{figure}

Let $f:\Omega\subset\R^d\to\R$ be Lipschitz. For $d=1$ and a level $u$, let
$N_u(f;\Omega)$ be the number of solutions of $f(x)=u$ in $\Omega$. For stationary
Gaussian processes, Rice's formula \eqref{eq:rice} gives the expected crossing rate as a
ratio of spectral moments: $\lambda_0$ is the signal variance and $\lambda_2$ the
derivative variance, so $\sqrt{\lambda_2/\lambda_0}$ is an RMS frequency. A field that
lacks high-frequency energy crosses every level less often
(Figure~\ref{fig:method}, right).

Two properties make the crossing density usable beyond the Gaussian setting. First, the
\emph{Kac--Rice integrand}: under mild conditions
\citep{azais2009level},
\begin{equation}
\E\,N_u \;=\; \int_\Omega \E\bigl[\,\lvert f'(x)\rvert \,\big|\, f(x)=u\,\bigr]\,
p_{f(x)}(u)\,dx ,
\end{equation}
which requires no stationarity. Second, and central for us, the deterministic
\emph{co-area formula} \citep{evans1992measure}: for Lipschitz $f$ and integrable $g$,
\begin{equation}
\int_\Omega g\bigl(f(x)\bigr)\,\lVert\nabla f(x)\rVert\,dx
\;=\;
\int_{\R} g(u)\,\mathcal{H}^{d-1}\bigl(f^{-1}(u)\cap\Omega\bigr)\,du,
\label{eq:coarea}
\end{equation}
where $\mathcal{H}^{d-1}$ is the $(d{-}1)$-dimensional Hausdorff measure. Taking
$g=\delta_\varepsilon(\cdot-u)$, a Gaussian kernel of bandwidth $\varepsilon$, the
left-hand side of \eqref{eq:coarea} divided by $\lvert\Omega\rvert$ becomes an
$\varepsilon$-smoothed \emph{level-set density}: crossings per unit length ($d{=}1$),
level-set length per unit area ($d{=}2$), level-set area per unit volume ($d{=}3$). No
probabilistic assumptions enter: \eqref{eq:coarea} holds for the single deterministic
field being fitted.

\subsection{A differentiable Monte-Carlo estimator}
\label{sec:estimator}

Sampling coordinates $x_1,\dots,x_N$ uniformly in $\Omega$ (or using the given training
coordinates) yields the estimator of \eqref{eq:estimator}:
$\hat c_\varepsilon(u)=\frac1N\sum_i
\delta_\varepsilon(f_\theta(x_i)-u)\,\lVert\nabla_x f_\theta(x_i)\rVert$ with
$\delta_\varepsilon(z)=(2\pi\varepsilon^2)^{-1/2}e^{-z^2/2\varepsilon^2}$.
It is unbiased for the $\varepsilon$-smoothed level-set density and inherits the usual
kernel trade-off: $O(\varepsilon^2)$ smoothing bias across levels versus
$O(1/(N\varepsilon))$ variance. Three implementation points matter in practice.

\emph{Gradients are free and exact.} $\nabla_x f_\theta$ comes from automatic
differentiation at arbitrary points; no finite differences, no grid. Backpropagating the
loss through $\lVert\nabla_x f_\theta\rVert$ is a double-backward operation, supported
by all standard frameworks \citep{paszke2019pytorch}.

\emph{Levels at target quantiles.} We place $L$ levels $u_1,\dots,u_L$ at uniformly
spaced quantiles of the ground-truth values in the batch (between the 2nd and 98th
percentile), which keeps every level populated regardless of the value histogram.

\emph{Same-batch targets.} The target profile $c^{\mathrm{gt}}(u_j)$ is the same
estimator applied to the ground-truth values and gradient norms \emph{at the same batch
points}. Estimator and target then share sampling noise, which partially cancels in the
difference---the loss compares two estimates of the same functional under the same
measure, so even under non-uniform sampling both sides are biased identically toward
densely sampled regions.

\subsection{The Kac--Rice loss}
\label{sec:loss}

The auxiliary loss matches the crossing-density profile at the $L$ levels, normalized to
be scale-free:
\begin{equation}
\loss_{\mathrm{KR}}(\theta)
\;=\;
\frac1L \sum_{j=1}^{L}
\frac{\bigl(\hat c_\varepsilon(u_j) - c^{\mathrm{gt}}(u_j)\bigr)^2}
     {\bigl(c^{\mathrm{gt}}(u_j) + \bar c^{\mathrm{gt}}\bigr)^2},
\qquad
\bar c^{\mathrm{gt}} = \tfrac1L\textstyle\sum_j c^{\mathrm{gt}}(u_j),
\label{eq:loss}
\end{equation}
and the total objective is
$\loss = \frac1N\sum_i (f_\theta(x_i)-y_i)^2 + \beta\,\loss_{\mathrm{KR}}$.
The normalization in \eqref{eq:loss} is not cosmetic: crossing densities grow linearly
with frequency content, so the \emph{squared} error grows quadratically, and an
unnormalized version couples the effective weight $\beta$ to the signal's bandwidth. In
early experiments the unnormalized loss with a weight tuned for one signal destabilized
training on another; the relative form transfers across signals with a single
$\beta$.\footnote{The same reasoning applies to the Sobolev baseline, which we normalize
by the mean squared ground-truth gradient norm; without this, its useful weight range
also varies by orders of magnitude across signals.}

Defaults throughout: $L=16$ levels, $\varepsilon=0.15\,\sigma_y$ where $\sigma_y$ is the
standard deviation of the batch ground-truth values, and $\beta=0.05$; all are ablated
in Section~\ref{sec:ablations}.

\paragraph{What the loss constrains.} $\loss_{\mathrm{KR}}$ is a functional of the
\emph{marginal} level-set geometry: it pushes the field to have the right amount of
level-crossing structure at each height, leaving \emph{where} that structure goes to the
reconstruction term. It cannot, alone, place an edge; it can insist that the total edge
length at each gray level be right. This is weaker than a Fourier loss on a grid (which
constrains amplitude at every frequency) but requires strictly less of the sampling
process, and it is robust to pointwise target noise by construction: a zero-mean error
of variance $\sigma^2$ on individual gradient-norm targets perturbs each
$c^{\mathrm{gt}}(u_j)$ only at $O(\sigma/\sqrt{N_j})$, with $N_j$ the effective number of
points near level $j$, whereas a pointwise gradient-matching loss absorbs that noise at
full variance.

\paragraph{Cost.} Relative to plain MSE, the additional cost is one backward pass for
$\nabla_x f_\theta$ (shared with any gradient-based loss) plus an $O(NL)$ kernel
evaluation, and double-backprop at optimization time. In a back-to-back measurement
(PE-MLP, $16{,}384$ points, CPU; \texttt{bench\_timing.py} in the released code, raw
output in the result files), one training iteration took $223$\,ms with MSE only,
$473$\,ms with the Kac--Rice term ($2.12\times$), and $479$\,ms with Sobolev
($2.14\times$)---the two gradient-based losses share the dominant double-backward
cost.

\section{Estimator validation}
\label{sec:validation}

\begin{figure}[t]
\centering
\includegraphics[width=.48\linewidth]{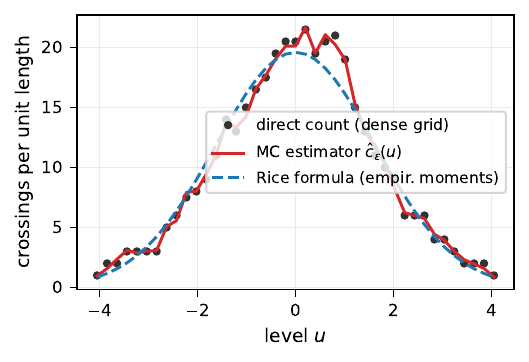}
\caption{\textbf{Estimator validation.} Smoothed Monte-Carlo estimator
$\hat c_\varepsilon(u)$ (red, $N{=}4\times10^5$ points, $\varepsilon{=}0.05$), exact
crossing counts from dense-grid sign changes (dots), and the Rice formula at empirical
spectral moments (dashed), for an approximately Gaussian random field (60 random-phase
sinusoids). The estimator tracks the exact counts across all levels, including the
non-Gaussian wiggles that the Rice formula idealizes away.}
\label{fig:validation}
\end{figure}

Before using \eqref{eq:estimator} inside a loss we verify it in isolation
(Figure~\ref{fig:validation}). On deterministic multisine signals the estimator matches
analytic crossing counts to within Monte-Carlo error (e.g., $\sin(\pi k x)$ on $[-1,1]$
yields measured density $k\pm2\%$ at $N=2\times10^5$, and the 2D co-area version
recovers the level-set length density of oriented gratings equally well). On an
approximately Gaussian random field (60 random-phase sinusoids), the estimator agrees
with direct sign-change counting on a dense grid within $5\%$ across levels
$u\in[-2.5\sigma, 2.5\sigma]$, and with the Rice prediction \eqref{eq:rice} evaluated at
empirical spectral moments within $10\%$. These tests ship as unit tests with the code.

\section{Experiments}
\label{sec:experiments}

\subsection{Protocol}
\label{sec:protocol}

\paragraph{Backbones.} The primary backbone for loss comparisons is a ReLU MLP on
Fourier positional encodings (PE-MLP; 3 hidden layers, width 256), the classic
spectrally biased INR \citep{tancik2020fourier}. SIREN \citep{sitzmann2020siren} and
FINER \citep{liu2024finer} appear as architecture baselines under plain MSE, and we test
composability by adding our loss to SIREN.

\paragraph{Baselines.} (i)~MSE only; (ii)~MSE + Focal Frequency Loss
\citep{jiang2021ffl}---our implementation is numerically verified against the official
package; (iii)~MSE + normalized Sobolev ($H^1$ gradient matching)
\citep{czarnecki2017sobolev,yuan2022sobolev}.

\paragraph{Tasks.} \emph{Regular grids} (sanity check): a five-tone multisine in 1D
(frequencies $2,5,11,23,47$ half-cycles per unit) fitted from 1{,}024 grid points, and
the \texttt{camera} image at $128^2$. \emph{Scattered domains} (the decisive test): the
$256^2$ \texttt{camera} image supervised only at $N{=}8{,}192$ scattered locations
(12.5\% of pixels) drawn from three densities---\texttt{uniform} (homogeneous),
\texttt{blobs} (75\% of samples in five Gaussian clusters), and \texttt{ramp} (linear
density gradient)---plus a 1D variant and a synthetic multi-scale texture. On scattered
tasks \emph{all} losses receive exactly the same information: the scattered pairs
$\{(x_i,y_i)\}$. FFL's grid target is obtained by Delaunay-based linear interpolation
(SciPy \texttt{griddata} \citep{virtanen2020scipy}) and the loss is applied on random
$128^2$ crops of that grid each iteration (patch-based FFL, a mode the official
implementation supports); Sobolev and Kac--Rice gradient targets are central differences
\emph{of that same interpolant} sampled at the training points. Oracle-gradient variants
(true image gradients; not legal supervision) isolate how much each gradient-based
method loses to interpolation noise.

\paragraph{Metrics.} PSNR on a dense evaluation grid; high-frequency PSNR (HF-PSNR:
PSNR of the residual after removing a Gaussian low-pass, $\sigma=4$\,px); SSIM
\citep{wang2004ssim}; and relative spectral error in eight radial frequency bands.
All results are mean $\pm$ s.d.\ over three seeds (model initialization and, on
scattered tasks, the sampling pattern). Adam \citep{kingma2015adam} with cosine decay;
full hyperparameters in Appendix~\ref{app:hyper}.

\subsection{Regular grids: parity, as expected}
\label{sec:exp1}

\begin{table}[t]
\centering
\caption{\textbf{Regular-grid fitting} (sanity check; mean $\pm$ s.d.\ over 3 seeds).
1D: parity across all losses. 2D: dense supervision needs no auxiliary drive---MSE-only
is best among PE-MLP losses, and architecture fixes (SIREN, FINER) dominate the entire
loss family on-grid.}
\label{tab:exp1}
\begin{tabular}{lcccc}
\toprule
 & \multicolumn{1}{c}{1D multisine} & \multicolumn{3}{c}{2D camera $128^2$} \\
\cmidrule(lr){2-2}\cmidrule(lr){3-5}
Config & PSNR & PSNR & HF-PSNR & SSIM \\
\midrule
MSE only & 70.64 $\pm$ 0.11 & 40.52 $\pm$ 1.10 & 42.46 $\pm$ 1.10 & 0.97 $\pm$ 0.00 \\
$+$ FFL & 69.72 $\pm$ 0.23 & 37.85 $\pm$ 0.52 & 39.80 $\pm$ 0.52 & 0.95 $\pm$ 0.00 \\
$+$ Sobolev & 70.41 $\pm$ 0.22 & 29.30 $\pm$ 0.12 & 31.26 $\pm$ 0.12 & 0.90 $\pm$ 0.00 \\
$+$ Kac-Rice (ours) & 70.28 $\pm$ 0.30 & 30.80 $\pm$ 1.00 & 32.79 $\pm$ 1.00 & 0.89 $\pm$ 0.02 \\
SIREN (MSE) & 93.77 $\pm$ 1.09 & 62.41 $\pm$ 0.33 & 64.33 $\pm$ 0.33 & 1.00 $\pm$ 0.00 \\
FINER (MSE) & 73.57 $\pm$ 7.68 & 139.72 $\pm$ 3.09 & 141.56 $\pm$ 3.04 & 1.00 $\pm$ 0.00 \\
\bottomrule
\end{tabular}

\end{table}

\begin{figure}[t]
\centering
\includegraphics[width=\linewidth]{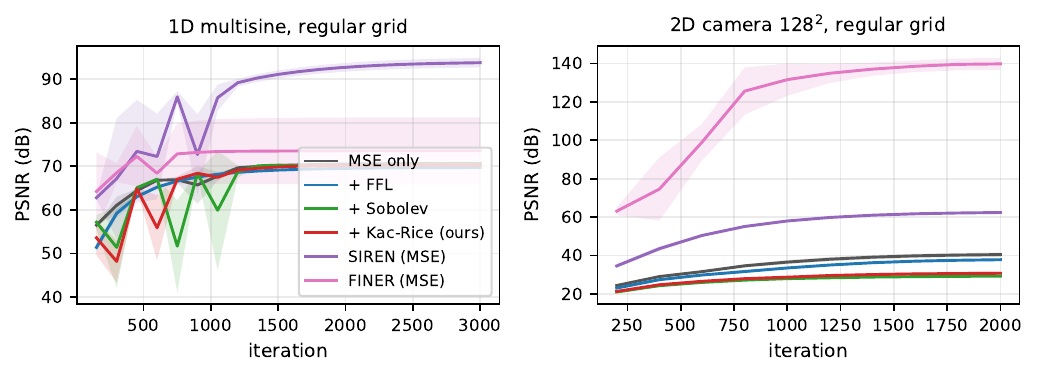}
\caption{\textbf{Regular-grid convergence.} PSNR versus iteration (mean over 3 seeds;
shaded $\pm$ s.d.). Left: 1D multisine---all PE-MLP losses converge to parity.
Right: $128^2$ image---with dense supervision, MSE-only is best and the
finite-difference gradient targets of Sobolev/Kac--Rice actively cost fidelity
(see text).}
\label{fig:exp1}
\end{figure}

\paragraph{1D: parity.} On the grid-sampled multisine, all four PE-MLP losses converge
to statistically indistinguishable quality (MSE $70.6\pm0.1$\,dB, FFL $69.7\pm0.2$,
Sobolev $70.4\pm0.2$, Kac--Rice $70.3\pm0.3$): with dense regular supervision, the
reconstruction term alone eventually recovers all five tones, and no auxiliary drive
changes the endpoint. SIREN, whose periodic activations match this signal family,
reaches $93.8\pm1.1$\,dB; FINER reaches $73.6\pm7.7$\,dB with high seed variance. This
is the expected sanity-check outcome: on-grid, auxiliary spectral losses neither help
nor hurt materially in 1D.

\paragraph{2D: dense supervision does not need---and is hurt by---auxiliary drives.}
On the $128^2$ image (Table~\ref{tab:exp1}), MSE-only training is the \emph{best}
configuration ($40.5\pm1.1$\,dB); FFL costs $2.7$\,dB, and the two gradient-based
losses cost ${\sim}10$\,dB. The mechanism for the latter is instructive: the gradient
targets are central differences of the discrete image, which are a low-pass-filtered
version of the true local differences the MSE term is driving toward. At high fidelity
the two objectives conflict, and a fixed $\beta$ biases the optimum away from exact
pixel reconstruction. (FFL's target is the exact pixel grid, so it remains consistent
with MSE and costs little.) \citet{yuan2022sobolev} similarly report that Sobolev
training for INRs requires care in how image derivatives are approximated. The
architecture baselines put the loss family in perspective: on-grid, SIREN reaches
$62.4\pm0.3$\,dB and FINER $139.7\pm3.1$\,dB (a numerically exact fit)---dense grid
fitting is simply solved by activation engineering, and no auxiliary loss on a PE-MLP
competes there. The practical reading, consistent with the few-shot literature
\citep{yang2023freenerf}: auxiliary spectral objectives are tools for
\emph{underdetermined} regimes, not for dense overfitting, and we evaluate them where
they are meant to operate---Section~\ref{sec:exp2}.

\subsection{Scattered non-uniform domains: the decisive test}
\label{sec:exp2}

\begin{table}[t]
\centering
\caption{\textbf{Scattered-domain fitting}: the decisive comparison ($256^2$
\texttt{camera}, $N{=}8{,}192$ samples, three sampling densities; mean $\pm$ s.d.\
over 3 seeds). All losses receive only the scattered samples. The \texttt{griddata}
row is the plain linear-interpolation reference computed from the same samples.}
\label{tab:exp2}
\resizebox{\textwidth}{!}{\begin{tabular}{lcccccc}
\toprule
 & \multicolumn{2}{c}{blobs} & \multicolumn{2}{c}{ramp} & \multicolumn{2}{c}{uniform} \\
\cmidrule(lr){2-3}\cmidrule(lr){4-5}\cmidrule(lr){6-7}
Config & PSNR & HF-PSNR & PSNR & HF-PSNR & PSNR & HF-PSNR \\
\midrule
griddata interp.\ (input) & 21.84 $\pm$ 0.10 & 25.27 $\pm$ 0.01 & 22.85 $\pm$ 0.10 & 25.70 $\pm$ 0.06 & 23.11 $\pm$ 0.10 & 25.84 $\pm$ 0.08 \\
\midrule
MSE only & 18.88 $\pm$ 0.12 & 22.76 $\pm$ 0.02 & 19.73 $\pm$ 0.09 & 22.89 $\pm$ 0.07 & 20.18 $\pm$ 0.07 & 23.19 $\pm$ 0.05 \\
$+$ FFL (interp.) & 21.62 $\pm$ 0.12 & 25.01 $\pm$ 0.05 & 22.48 $\pm$ 0.12 & 25.31 $\pm$ 0.09 & 22.73 $\pm$ 0.11 & 25.46 $\pm$ 0.09 \\
$+$ Sobolev (est. grad) & 21.59 $\pm$ 0.15 & 25.07 $\pm$ 0.09 & 22.74 $\pm$ 0.07 & 25.59 $\pm$ 0.05 & 23.02 $\pm$ 0.08 & 25.75 $\pm$ 0.05 \\
$+$ Kac-Rice (est. grad, ours) & 21.14 $\pm$ 0.17 & 24.71 $\pm$ 0.07 & 22.31 $\pm$ 0.15 & 25.25 $\pm$ 0.12 & 22.53 $\pm$ 0.15 & 25.36 $\pm$ 0.13 \\
$+$ Sobolev (oracle grad) & 21.80 $\pm$ 0.21 & 25.36 $\pm$ 0.13 & -- & -- & -- & -- \\
$+$ Kac-Rice (oracle grad) & 20.86 $\pm$ 0.20 & 24.50 $\pm$ 0.08 & -- & -- & -- & -- \\
SIREN (MSE) & 20.10 $\pm$ 0.20 & 23.97 $\pm$ 0.15 & -- & -- & -- & -- \\
SIREN $+$ FFL (interp.) & 21.92 $\pm$ 0.11 & 25.38 $\pm$ 0.03 & -- & -- & -- & -- \\
SIREN $+$ Sobolev (est. grad) & 21.50 $\pm$ 0.14 & 25.06 $\pm$ 0.10 & -- & -- & -- & -- \\
SIREN $+$ Kac-Rice (ours) & 21.59 $\pm$ 0.10 & 25.14 $\pm$ 0.05 & -- & -- & -- & -- \\
\bottomrule
\end{tabular}
}
\end{table}

\begin{figure}[t]
\centering
\includegraphics[width=\linewidth]{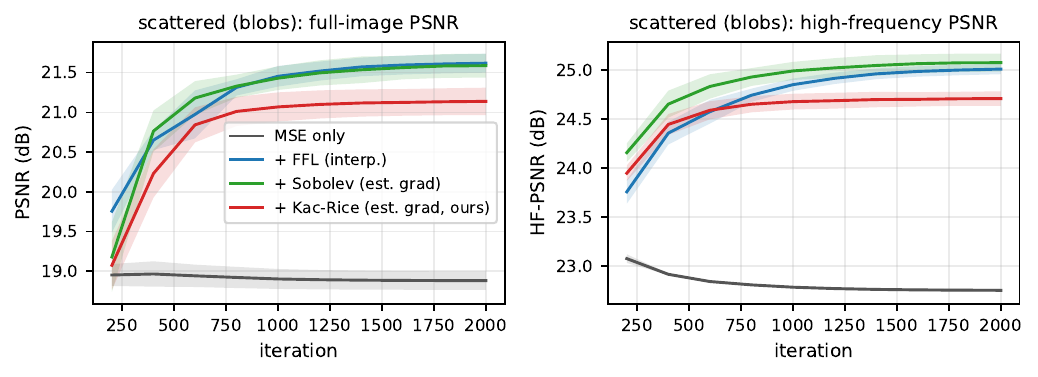}
\caption{\textbf{Scattered-domain convergence} (\texttt{blobs} sampling; mean over 3
seeds, shaded $\pm$ s.d.). Full-image PSNR (left) and high-frequency PSNR (right).
MSE-only stalls immediately; all three auxiliary losses climb $2.3$--$2.7$\,dB above
it. Kac--Rice leads on HF-PSNR early (its target does not need to warm up any grid
structure) before Sobolev and FFL edge ahead after ${\sim}700$ iterations.}
\label{fig:exp2curves}
\end{figure}

\begin{figure}[t]
\centering
\includegraphics[width=\linewidth]{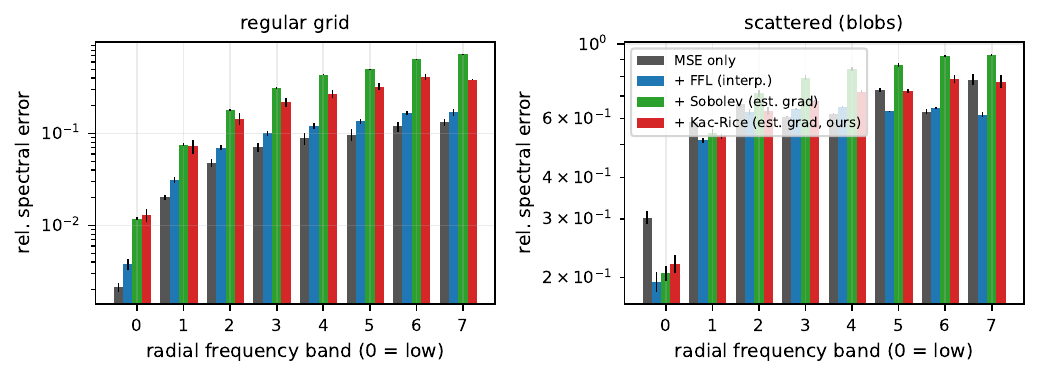}
\caption{\textbf{Per-band relative spectral error} (log scale; lower is better;
mean $\pm$ s.d.\ over 3 seeds). Left, regular grid: MSE-only is lowest in every band,
mirroring Table~\ref{tab:exp1}. Right, scattered \texttt{blobs}: the auxiliary losses
dominate MSE in the low bands, and in the two highest bands Kac--Rice attains the
lowest error among the spatial-domain losses, consistent with its role as a
high-frequency drive.}
\label{fig:bands}
\end{figure}

\begin{figure}[t]
\centering
\includegraphics[width=\linewidth]{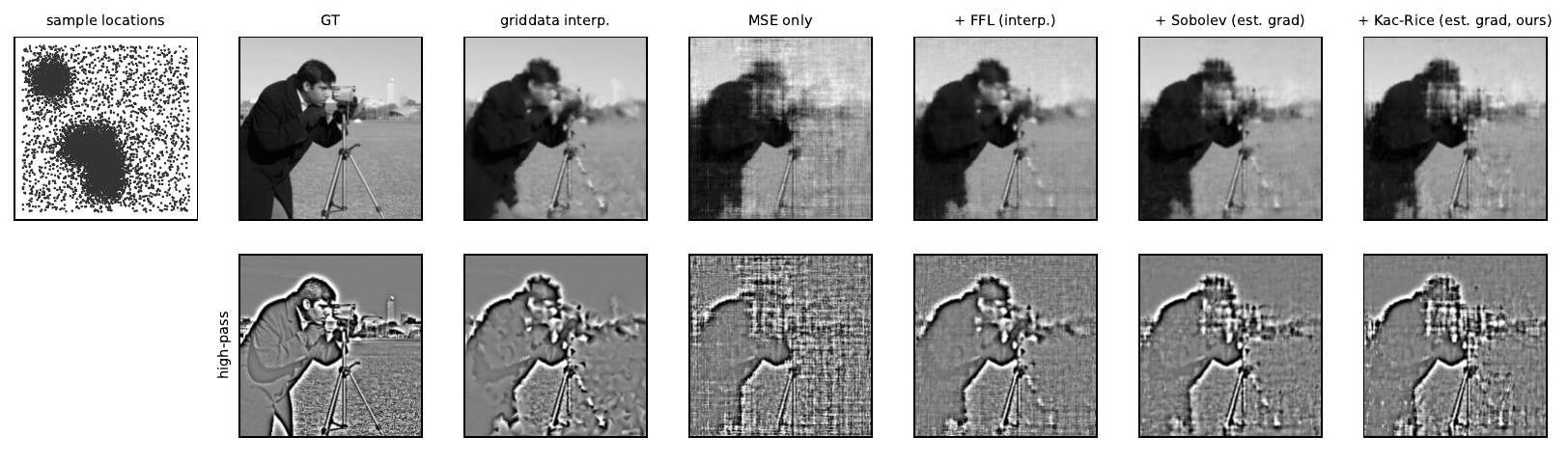}
\caption{\textbf{Qualitative results} on scattered \texttt{blobs} sampling (seed 0):
sample locations, reconstructions (top), high-pass residuals (bottom). MSE-only
produces streaked, washed-out structure; every auxiliary loss restores coherent
detail. Differences among the three auxiliary losses are subtle at this scale---%
consistent with the ${\le}0.5$\,dB spreads in Table~\ref{tab:exp2}.}
\label{fig:recons}
\end{figure}

\paragraph{All auxiliary losses matter here; Kac--Rice ties FFL but does not beat it.}
Table~\ref{tab:exp2} is the decisive comparison, and we report it as measured. First,
the regime change from Section~\ref{sec:exp1} is dramatic: with scattered supervision,
\emph{every} auxiliary loss now adds $+2.3$ to $+3.0$\,dB over MSE-only training
(e.g.\ \texttt{blobs}: $18.9\to21.1$--$21.6$\,dB), and the qualitative gap is larger
than the numbers suggest---the MSE-only reconstructions in Figure~\ref{fig:recons} are
blurred and streaked, while all three auxiliary variants recover coherent structure.
Second, the central hypothesis of this paper is \emph{not} confirmed on the PE-MLP
backbone: Kac--Rice lands within $0.2$--$0.5$\,dB of FFL-with-resampling on every
sampling density (\texttt{blobs} $21.1$ vs.\ $21.6$; \texttt{ramp} $22.3$ vs.\ $22.5$;
\texttt{uniform} $22.5$ vs.\ $22.7$) but does not overtake it, and the normalized
Sobolev loss is the strongest or tied-strongest loss throughout. On SSIM the ordering
partially reverses---Kac--Rice matches FFL on \texttt{blobs} and exceeds it on
\texttt{ramp} and \texttt{uniform} ($0.60/0.66/0.66$ vs.\ $0.59/0.60/0.62$)---%
indicating the distributional loss buys structural coherence rather than pixel-wise
accuracy; per-band spectra (Figure~\ref{fig:bands}, right) agree,
with Kac--Rice showing the lowest error of the spatial-domain losses in the top two
frequency bands. But the headline claim we set out to test---a clear PSNR win for
crossing densities off-grid---did not materialize.

\paragraph{The interpolation ceiling.} The \texttt{griddata} row of
Table~\ref{tab:exp2} is sobering context: at this sample budget, plain linear
interpolation of the scattered samples scores within noise of the \emph{best} INR
configuration on PSNR ($21.8$ on \texttt{blobs}). On the natural image, no loss
studied here---frequency-domain or spatial---pushes a PE-MLP past the information
ceiling that the interpolant itself sets (the texture task below is the sole,
marginal exception). What the INR adds over \texttt{griddata} is a compact,
continuous, differentiable representation, not extra pixels; auxiliary losses determine
how close to the ceiling one gets.

\paragraph{Content dependence: crossing statistics win on texture.}
The synthetic multi-scale texture (Table~\ref{tab:synthetic}) reverses the ordering.
There Kac--Rice scores $27.19\pm0.33$\,dB (SSIM $0.79$) against FFL's $26.58\pm0.15$
(SSIM $0.74$)---a $+0.6$\,dB advantage---with Sobolev at $27.29\pm0.13$; the two
spatial-domain losses match or slightly exceed the \texttt{griddata} ceiling
($26.96$) while FFL stays below it. The pattern is theoretically coherent: the
texture is statistically homogeneous, i.e.\ close to the stationary-random-field
regime in which crossing densities are an exact spectral functional
\eqref{eq:rice}, whereas \texttt{camera} is edge-dominated and non-stationary, so its
crossing profile is a blunter summary. Crossing-density supervision is at its best
when the signal looks like the random fields the theory was built for.

\paragraph{Oracle diagnostics: the distributional loss is robust to target smoothing.}
Replacing estimated gradients with oracle image gradients \emph{helps} Sobolev
($21.6\to21.8$\,dB) but does not help Kac--Rice ($21.1\to20.9$\,dB). This is the
clearest direct evidence for the robustness mechanism of
Section~\ref{sec:loss}: Sobolev consumes gradient targets pointwise and benefits from
their exactness, whereas Kac--Rice only consumes their batch statistics---smoothing
noise largely cancels, and sharper targets, if anything, ask for crossing mass the
MSE anchor cannot yet place. The practical consequence: Kac--Rice's performance is
insensitive to the quality of the gradient-estimation step, which is the step that is
genuinely hard on scattered domains.

\subsection{Composability with SIREN}
\label{sec:composability}

Because the loss is orthogonal to architecture, it should stack with activation-based
remedies. On the \texttt{blobs} task, SIREN+MSE scores $20.10\pm0.20$\,dB (SSIM
$0.56$); adding Kac--Rice lifts it to $21.59\pm0.10$ (SSIM $0.65$). To test whether
this is specific to our loss we also ran SIREN with the two baseline auxiliaries: FFL
reaches $21.92\pm0.11$ (SSIM $0.69$) and Sobolev $21.50\pm0.14$ (SSIM $0.63$). The
honest conclusion is that \emph{composability is a property of auxiliary spectral
supervision in general}, not of crossing densities in particular---though two details
favor the distributional loss here: on SIREN, Kac--Rice edges Sobolev (the reverse of
their PE-MLP ordering), and every SIREN+auxiliary configuration beats its PE-MLP
counterpart, so the loss and the activation fix address genuinely different parts of
the problem.

\subsection{Ablations}
\label{sec:ablations}

\begin{figure}[t]
\centering
\includegraphics[width=\linewidth]{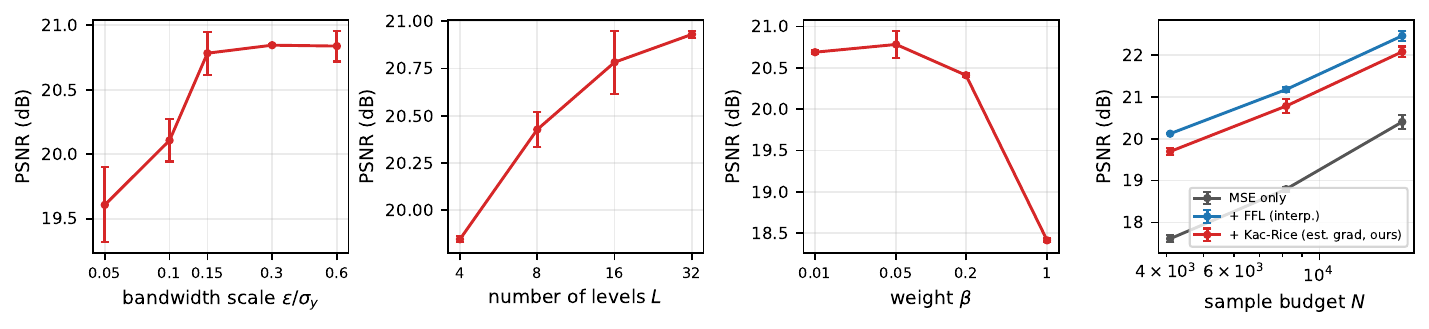}
\caption{\textbf{Ablations} on \texttt{camera}/\texttt{blobs} (1{,}000 iterations, 2
seeds; error bars $\pm$ s.d.). Left to right: Dirac bandwidth scale
$\varepsilon/\sigma_y$ (wide plateau above $0.15$), number of levels $L$ (saturates by
$16$--$32$), loss weight $\beta$ (flat over a $20\times$ range, collapsing only at
$\beta{=}1$), and sample budget $N$ (orderings stable across budgets).}
\label{fig:ablations}
\end{figure}

All ablations use \texttt{camera}/\texttt{blobs} at 1{,}000 iterations with 2 seeds
(Figure~\ref{fig:ablations}); absolute numbers are therefore lower than in
Table~\ref{tab:exp2}, and only relative sensitivity is meaningful.

\paragraph{Bandwidth $\varepsilon$.} The loss is robust for
$\varepsilon\ge0.15\,\sigma_y$ (PSNR $20.8$--$20.9$\,dB across
$\varepsilon/\sigma_y\in\{0.15,0.3,0.6\}$) and degrades below it
($19.6$\,dB at $0.05$): a too-narrow Dirac starves most batch points of gradient
signal, exactly the failure mode anticipated. The practical guidance is simply
``not too small''---the nuisance parameter flagged at the outset turns out to have a
wide plateau rather than a narrow sweet spot.

\paragraph{Number of levels $L$.} Quality improves monotonically and saturates:
$19.9$, $20.4$, $20.8$, $20.9$\,dB for $L=4,8,16,32$. Sixteen levels suffice.

\paragraph{Weight $\beta$.} Performance is flat across $\beta\in[0.01,0.2]$
($20.4$--$20.8$\,dB) and collapses only at $\beta=1$ ($18.4$\,dB), where the
distributional term overwhelms reconstruction. The relative normalization of
\eqref{eq:loss} is what makes this range transferable across signals.

\paragraph{Sample budget.} Across $N\in\{4096, 8192, 16384\}$ the ordering is stable,
with FFL ahead of Kac--Rice by a constant ${\sim}0.4$\,dB ($0.42/0.40/0.38$). The
auxiliary advantage over MSE is large at every budget but contracts slowly as
sampling densifies---from $+2.5/+2.1$\,dB (FFL/Kac--Rice) at $N{=}4096$ to
$+2.1/+1.7$\,dB at $N{=}16384$---consistent with auxiliary drives mattering most
when supervision is scarcest.

\section{Discussion and limitations}
\label{sec:discussion}

\paragraph{What we learned.} Taking the experiments together, the honest verdict on
the opening hypothesis is \emph{parity with a content-dependent edge, not a general
victory}: the crossing-density loss is a working, validated, mesh-free high-frequency
drive that matches frequency-domain and gradient-matching supervision on
edge-dominated natural-image content, beats MSE-only training by large margins
wherever supervision is scarce, and overtakes FFL only where the signal is
statistically homogeneous (texture)---the regime the underlying theory is actually
about. Its remaining distinguishing properties are structural. First, it is the only loss in the comparison that needs
neither a grid at any point in the pipeline (FFL needs one for its transform) nor
trustworthy pointwise gradient targets (Sobolev's accuracy tracks target quality,
while Kac--Rice is insensitive to it---the oracle diagnostic). Second, it buys
structural metrics preferentially: SSIM matching (\texttt{blobs}) or exceeding
(\texttt{ramp}, \texttt{uniform}) FFL's, and the best top-band spectral error among
the spatial-domain losses, at a small PSNR cost. Where those structural properties are decisive---supervision without any usable
grid, gradient targets too noisy to match pointwise---it is a reasonable default;
where a grid or clean gradients exist, existing losses are already sufficient.

\paragraph{Why no win?\ } The mechanism we can defend from the data: the crossing
profile constrains $L$ numbers per batch---a far weaker constraint than FFL's full
spectrum or Sobolev's $N$ pointwise gradients. Robustness and weakness here are two
faces of the same statistic. The interpolation-ceiling observation sharpens this: at
$N{=}8{,}192$ samples all auxiliary losses already saturate what linear interpolation
extracts from the samples, so there was little headroom for a cleverer loss to
express an advantage. Regimes where the ceiling itself moves---strong shape priors,
composed losses, or targets (b)-style scheduled crossing profiles that inject
\emph{prior} spectral knowledge rather than re-encoding sample information---are where
a distributional drive could still separate; all remain untested here.

\paragraph{Limitations.} (i)~Scope: 1D signals and 2D images only; neural SDFs from
point clouds---arguably the setting where ``no grid, noisy normals'' bites hardest---%
and few-shot NeRF remain future work, and our conclusions should not be extrapolated
to them. (ii)~The loss constrains distributional level-set geometry, not phase: it
cannot place an edge, only demand total edge mass, so it is strictly an auxiliary
term. (iii)~Hyperparameters are benign but real: $\varepsilon$ has a wide plateau
above $0.15\sigma_y$, $L{=}16$ suffices, and $\beta$ tolerates a $20\times$ range---%
but all were tuned on the same image family used for evaluation. (iv)~Double backprop
makes each iteration ${\sim}2.1\times$ an MSE iteration, matching Sobolev.
(v)~Evaluation uses one natural test image plus one synthetic texture at one
resolution per task; broader image suites would tighten the estimates, though the
seed-level spreads are already small relative to the effects discussed.
(vi)~\texttt{camera} is a standard public test image; no personal or sensitive data
is involved.

\section{Conclusion}
\label{sec:conclusion}

We turned the Rice level-crossing density into a differentiable, mesh-free,
FFT-free auxiliary loss for INRs, validated the underlying estimator against exact
crossing counts and the Rice formula, and evaluated it against the strongest loss-based
alternatives under strict information parity. The result is a clean, honest data
point for the field: crossing statistics \emph{work} as a high-frequency drive---%
recovering $+2.3$--$3.0$\,dB over plain reconstruction on PE-MLP and
$+1.4$--$1.8$\,dB on SIREN wherever supervision is scarce---but on natural images they
match rather than beat
frequency-domain and Sobolev supervision, pulling ahead of the frequency-domain loss
only on statistically homogeneous texture, with their broader distinction being
structural: no grid, no FFT, and demonstrated insensitivity to gradient-target noise. The broader point is
that the geometry of level sets offers a family of spatial-domain spectral
surrogates---crossing densities here; critical-point counts and Euler characteristics
of excursion sets \citep{adler2007random} are natural sequels---that remain defined
exactly where frequency-domain tooling is not, and the settings that motivated this
loss most strongly (point-cloud SDFs, scheduled crossing targets) remain open.

\paragraph{Reproducibility.} Code, experiment scripts, raw result JSONs, and the
estimator's unit tests are public at
\url{https://github.com/gunnerhowe/Research} (directory \texttt{kac-rice}). All
experiments were run on a single consumer workstation (16-core CPU / RTX~3080); the
complete suite reproduces in under a day of commodity compute.

\bibliographystyle{plainnat}
\bibliography{references}

\appendix

\section{Hyperparameters}
\label{app:hyper}

\begin{table}[h]
\centering
\caption{Hyperparameters for all experiments.}
\begin{tabular}{ll}
\toprule
Backbone (loss comparisons) & PE-MLP: ReLU, 3 hidden layers, width 256 \\
Positional encoding & $\{\sin,\cos\}(2^k\pi x)$, $k=0..5$ (1D), $k=0..7$ (2D) \\
SIREN / FINER & width 256, 3 hidden layers, $\omega_0=30$; FINER default init \\
Optimizer & Adam, lr $10^{-3}$ (PE-MLP) / $5\times10^{-4}$ (SIREN, FINER) \\
Schedule & cosine decay to $0.05\times$ lr \\
Iterations & 3{,}000 (1D), 2{,}000 (2D), 1{,}000 (ablations); full batch \\
Kac--Rice loss & $L=16$ levels at quantiles $[0.02,0.98]$; $\varepsilon=0.15\sigma_y$; $\beta=0.05$ \\
Sobolev loss & normalized $H^1$; weight $0.05$ \\
FFL & $\alpha=1$, weight $1.0$; random $128^2$ crops on scattered tasks \\
Scattered supervision & $N=8{,}192$ samples of the $256^2$ image (12.5\%) \\
Seeds & 3 (init + sampling pattern); 2 for ablations \\
\bottomrule
\end{tabular}
\end{table}

\section{Reproducibility}
\label{app:repro}

All code, raw result JSONs, and figure generators are public at
\url{https://github.com/gunnerhowe/Research} (directory \texttt{kac-rice}).
Every number and figure in this paper regenerates from the released raw runs;
no result exists only in the manuscript.

\paragraph{Environment.} Python 3.13.6, PyTorch 2.7.1, NumPy 2.4.4, SciPy
1.17.0, Matplotlib 3.10.7, on a single 16-core Windows workstation. All
headline experiments ran on CPU in float32 (the machine's GPU was shared with
unrelated jobs; CPU wall-clock is not a reported metric, and the timing
comparison of Section~3 uses the dedicated back-to-back benchmark below).

\paragraph{Commands.} Estimator correctness tests:
\texttt{python tests/test\_correctness.py} (7 tests: crossing counts, co-area,
Rice agreement, FFL parity vs.\ the official package, gradient flow). Full
experiment suite (parallel workers; resume-safe, so a single sequential
invocation is equivalent):
\begin{center}\small\ttfamily
python experiments/run\_paper.py --suites exp1\_1d exp2\_1d exp1\_2d\\
python experiments/run\_paper.py --suites exp2\_2d --tag \_a\\
python experiments/run\_paper.py --suites ablations --tag \_c
\end{center}
Figures and tables: \texttt{python experiments/make\_figures.py} reads
\texttt{results/paper/*.json} and writes every figure and LaTeX table used
here. Timing artifact: \texttt{python experiments/bench\_timing.py} writes
\texttt{results/paper/timing.json} (source of the $2.12\times$/$2.14\times$
factors).

\paragraph{Seeds and scope.} Seeds $0$--$2$ everywhere (model init and, on
scattered tasks, sampling patterns), $0$--$1$ for ablations; all seeds are set
by CLI and recorded per run in the JSONs. Total compute: under one day on the
single workstation.

\section{Additional results}
\label{app:additional}

\begin{figure}[h]
\centering
\includegraphics[width=\linewidth]{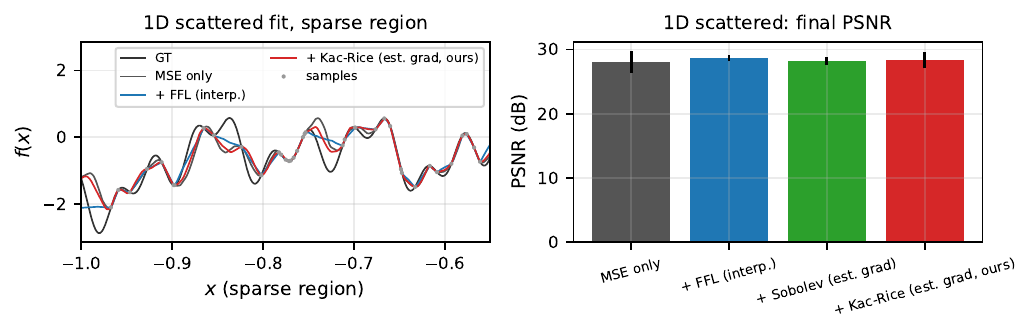}
\caption{\textbf{1D scattered fitting} (ramp density, $N{=}384$ samples,
mean over 3 seeds). Left: reconstructions in the sparsely sampled region; all methods
undershoot the highest tone where samples are scarce, with FFL producing the smoothest
(most conservative) fit. Right: final PSNR---all four configurations tie within noise
($28.1$--$28.6$\,dB). In 1D the linear interpolant is already a strong gradient/grid
surrogate, and no auxiliary loss separates from the pack.}
\label{fig:exp2_1d}
\end{figure}

\begin{table}[h]
\centering
\caption{Scattered \texttt{blobs} fitting on the synthetic multi-scale texture
($256^2$, $N{=}8{,}192$; mean $\pm$ s.d.\ over 3 seeds).}
\label{tab:synthetic}
\begin{tabular}{lccc}
\toprule
Config & PSNR & HF-PSNR & SSIM \\
\midrule
MSE only & 24.81 $\pm$ 0.88 & 19.78 $\pm$ 0.80 & 0.59 $\pm$ 0.06 \\
$+$ FFL (interp.) & 26.58 $\pm$ 0.15 & 21.64 $\pm$ 0.15 & 0.74 $\pm$ 0.01 \\
$+$ Sobolev (est. grad) & 27.29 $\pm$ 0.13 & 22.23 $\pm$ 0.22 & 0.81 $\pm$ 0.01 \\
$+$ Kac-Rice (est. grad, ours) & 27.19 $\pm$ 0.33 & 22.17 $\pm$ 0.10 & 0.79 $\pm$ 0.02 \\
\bottomrule
\end{tabular}

\end{table}

\end{document}